\documentclass[11pt]{article}
\usepackage[margin=1in]{geometry}
\usepackage{setspace}
\usepackage{amsmath,amssymb,amsthm}
\usepackage{mathtools}
\usepackage{graphicx}
\usepackage{booktabs}
\usepackage{multirow}
\usepackage{array}
\usepackage[round,sort]{natbib}
\usepackage[colorlinks=true,linkcolor=blue,citecolor=blue,urlcolor=blue]{hyperref}
\usepackage{algorithm}
\usepackage{algorithmic}
\usepackage{enumitem}
\usepackage{xcolor}
\usepackage{tikz}
\usetikzlibrary{decorations.pathreplacing,calc,arrows.meta,shapes.geometric}

\title{Distributed Training using an Intelligent Network}
\author{
  Nihar Shah\\
  DoubleZero Foundation\\
  \texttt{nihar@doublezero.us}
  \and
  Ben Blier\\
  DoubleZero Foundation\\
  \texttt{ben@doublezero.us}
}
\date{\vspace{3mm}\today}

\begin{document}
\maketitle

\begin{abstract}
Distributed training across a wide area network (WAN) is challenging, as continuous parameter exchange by islands of compute is constrained by limited bandwidth, high latency, and uneven topology. We propose making the network an active participant in training. On the systems side, such networks should leverage (i) multicast technology to replicate outbound traffic and (ii) in-line FPGAs to aggregate inbound traffic, to ease egress and ingress bottlenecks. These technologies are used for training across workers within a data center, but this paper extends them to the WAN. On the algorithms side, we develop an optimization framework that produces rich synchronization schedules (namely, rotating cliques of islands) around the underlying network topology and these technologies, to maximize information exchange. Finally, we illustrate this on a nine-city topology modeled on the DoubleZero network, a live programmable WAN equipped with both technologies, and show how the optimal schedules shift with the network's capabilities. Together, these can narrow the gap to the gold standard of colocated training.
\end{abstract}

\newpage
\section{Introduction}
\label{sec:introduction}
Model training is increasingly outgrowing any individual data center. Power and permitting are strong limiting forces \citep{patel2024multidatacenter}, and in some cases valuable training data cannot leave a given location for compliance reasons. Distributed training across multiple data centers is thus becoming more commonplace.\footnote{As two examples: Google has disclosed that Gemini was trained across multiple data centers \citep{geminiteam2023gemini}, and INTELLECT-1 trained a ten-billion parameter model across three continents \citep{jaghouar2024intellect1}.}

For distributed training to be successful, it should stay as close to synchronous SGD as possible. More specifically, parallel workers (which each hold a copy of the model) compute gradients on local data batches, aggregate them, and then update the model before proceeding to the next batch \citep{goyal2017accurate}. Within a data center, where connectivity is fast, symmetric, and plentiful, this is generally feasible.

But when workers are spread over a wide area network (WAN), this becomes infeasible. Islands of compute, which hold groups of colocated workers (to follow the terminology of \citet{douillard2023diloco}), are connected to one another over long, low-capacity, and asymmetric links. This means the synchronization step becomes the bottleneck, as a naive application of synchronous SGD would require far too much idling of the individual workers. Much of the existing literature thus adapts the training around the network, e.g. synchronizing rarely, compressing aggressively, or tolerating staleness.

Our paper makes the network an active participant in the training, by proposing two components. This proposal is described below, and visualized across a stylized seven-island setup in Figure \ref{fig:rotating}.
\begin{enumerate}
\item The first is a systems proposal. We propose using multicast technology and FPGAs across the network to perform replication and aggregation of model updates from islands of compute. Multicast conserves bandwidth (especially egress bandwidth from islands) by doing just-in-time replication; and FPGAs conserve ingress bandwidth to islands by aggregating inbound streams at the WAN edge. Both operate at the line rate of data, adding no serialization delay. This scales more efficiently to many islands without requiring parameter servers or other hierarchical systems, which add their own latency and bandwidth considerations.
\item The second is an algorithms proposal, which is an optimization framework to set a synchronization schedule between islands. Most existing methodologies do not interact with the network topology or do so only lightly; but we construct rich schedules of rotating cliques of islands directly from the network's topology and technological capabilities. This framework trades off information transmission against time, and finds the most efficient point.
\end{enumerate}

\begin{figure}[htbp]
\centering
\resizebox{0.95\textwidth}{!}{%
\begin{tikzpicture}[
  isl/.style={circle, draw=black, minimum size=15pt, inner sep=1pt, font=\scriptsize},
  islidle/.style={isl, fill=white, densely dotted},
  hub/.style={circle, draw=black, fill=gray!40, minimum size=9pt, inner sep=0pt},
  fpga/.style={rectangle, draw=black, fill=white, minimum size=6pt, inner sep=0pt, line width=0.7pt},
  core/.style={line width=2.4pt, draw=gray!45},
  access/.style={line width=0.5pt, draw=gray!85},
  flow/.style={-{Stealth[length=2.5mm]}, line width=1.0pt, draw=black},
  tie/.style={densely dotted, line width=0.7pt, draw=black},
  note/.style={font=\scriptsize, align=left}
]
\begin{scope}
  \node[font=\bfseries\small, anchor=west] at (-0.8,7.8)
    {Round 1: two triangles \; {\normalfont\small (island 5 sits out)}};
  \node[hub] (A) at (1.2,4.6) {}; \node[hub] (B) at (4.0,5.3) {};
  \node[hub] (C) at (6.6,4.2) {}; \node[hub] (D) at (4.8,2.2) {};
  \node[hub] (E) at (1.6,1.6) {};
  \draw[core] (A)--(B); \draw[core] (B)--(C); \draw[core] (C)--(D);
  \draw[core] (D)--(E); \draw[core] (E)--(A); \draw[core] (B)--(D);
  \node[isl, fill=gray!15] (I1) at (0.1,5.7) {1};
  \node[isl, fill=gray!50] (I2) at (0.2,3.6) {2};
  \node[isl, fill=gray!15] (I3) at (4.7,6.4) {3};
  \node[isl, fill=gray!15] (I4) at (7.7,5.1) {4};
  \node[islidle]           (I5) at (7.5,3.1) {5};
  \node[isl, fill=gray!50] (I6) at (5.9,1.0) {6};
  \node[isl, fill=gray!50] (I7) at (0.5,0.5) {7};
  \foreach \i/\h in {I1/A,I2/A,I3/B,I4/C,I5/C,I6/D,I7/E}{
    \draw[access] (\i) -- (\h);
    \node[fpga] at ($(\i)!0.5!(\h)$) {};
  }
  \draw[tie] (I1) to[bend left=20]  (I3);
  \draw[tie] (I3) to[bend left=15]  (I4);
  \draw[tie] (I4) to[bend left=28] (I1);
  \draw[tie] (I2) to[bend right=14] (I7);
  \draw[tie] (I7) to[bend right=15] (I6);
  \draw[tie] (I6) to[bend left=12]  (I2);
  \node[note, anchor=west] at ($(I5)+(0.35,-0.4)$) {(sits out)};
\end{scope}
\begin{scope}[xshift=10.2cm]
  \node[font=\bfseries\small, anchor=west] at (-0.8,7.8)
    {Round 2: a triangle and two pairs};
  \node[hub] (A2) at (1.2,4.6) {}; \node[hub] (B2) at (4.0,5.3) {};
  \node[hub] (C2) at (6.6,4.2) {}; \node[hub] (D2) at (4.8,2.2) {};
  \node[hub] (E2) at (1.6,1.6) {};
  \draw[core] (A2)--(B2); \draw[core] (B2)--(C2); \draw[core] (C2)--(D2);
  \draw[core] (D2)--(E2); \draw[core] (E2)--(A2); \draw[core] (B2)--(D2);
  \node[isl, fill=gray!15] (J1) at (0.1,5.7) {1};
  \node[isl, fill=gray!50] (J2) at (0.2,3.6) {2};
  \node[isl, fill=gray!50] (J3) at (4.7,6.4) {3};
  \node[isl, fill=gray!15] (J4) at (7.7,5.1) {4};
  \node[isl, fill=gray!80, text=white] (J5) at (7.5,3.1) {5};
  \node[isl, fill=gray!80, text=white] (J6) at (5.9,1.0) {6};
  \node[isl, fill=gray!15] (J7) at (0.5,0.5) {7};
  \foreach \i/\h in {J1/A2,J2/A2,J3/B2,J4/C2,J5/C2,J6/D2,J7/E2}{
    \draw[access] (\i) -- (\h);
    \node[fpga] at ($(\i)!0.5!(\h)$) {};
  }
  \draw[tie] (J1) to[bend left=22] (J4);
  \draw[tie] (J4) to[bend left=14] (J7);
  \draw[tie] (J7) to[bend left=10] (J1);
  \draw[tie] (J2) to[bend left=26] (J3);
  \draw[tie] (J5) to[bend left=20] (J6);
\end{scope}
\begin{scope}[yshift=-5.6cm, xshift=2.6cm]
  \node[font=\bfseries\small, anchor=west] at (-1.6,3.5)
    {Zoom into the \{1,4,7\} clique's streams in round 2};
  \node[hub] (P) at (2.8,1.9) {};
  \node[hub] (Q) at (2.8,-0.7) {};
  \node[hub] (M) at (5.6,0.6) {};
  \node[hub] (R) at (8.6,0.6) {};
  \draw[core] (P)--(M); \draw[core] (Q)--(M); \draw[core] (M)--(R);
  \node[isl, fill=gray!15] (z1) at (0.8,2.4) {1};
  \node[isl, fill=gray!15] (z7) at (0.8,-1.2) {7};
  \node[isl, fill=gray!15] (z4) at (11.8,0.6) {4};
  \draw[access] (z1)--(P); \draw[access] (z7)--(Q); \draw[access] (R)--(z4);
  \node[fpga] at ($(z1)!0.5!(P)$) {};
  \node[fpga] at ($(z7)!0.5!(Q)$) {};
  \node[fpga] (f4) at ($(R)!0.5!(z4)$) {};
  \draw[flow] (z1) to[bend left=18] (P);
  \draw[flow] (P) to[bend left=12] (M);
  \draw[flow] (z7) to[bend right=18] (Q);
  \draw[flow] (Q) to[bend right=12] (M);
  \draw[flow] (P) -- (2.0,1.15);
  \node[note, anchor=east] at (1.9,1.15) {(copy to 7)};
  \draw[flow] (Q) -- (2.0,0.05);
  \node[note, anchor=east] at (1.9,0.05) {(copy to 1)};
  \draw[flow] ($(M)+(0.12,0.14)$) -- ($(R)+(-0.14,0.14)$);
  \draw[flow] ($(M)+(0.12,-0.14)$) -- ($(R)+(-0.14,-0.14)$);
  \draw[flow] ($(R)+(0.14,0.10)$) -- ($(f4.west)+(-0.03,0.05)$);
  \draw[flow] ($(R)+(0.14,-0.10)$) -- ($(f4.west)+(-0.03,-0.05)$);
  \draw[flow, line width=1.5pt] (f4) -- (z4);
  \node[note, anchor=east]  at (0.35,-1.2) {(i) one stream out\\of each sender};
  \node[note, anchor=south] at (7.1,0.95)  {(ii) side by side on the core};
  \node[note, anchor=north] at (10.2,0.3) {(iii) aggregated at\\island 4's edge FPGA};
  \node[note, anchor=south] at (11.5,0.95) {(iv) one stream in};
\end{scope}
\begin{scope}[yshift=-9.0cm, xshift=0.4cm]
  \node[isl, fill=gray!50, minimum size=11pt] at (0.0,0) {0};
  \node[note, anchor=west] at (0.3,0) {island};
  \node[hub] at (4.6,0) {};
  \node[note, anchor=west] at (4.9,0) {hub};
  \draw[access] (6.6,0) -- (7.2,0);
  \node[note, anchor=west] at (7.35,0) {thin access link};
  \draw[core] (10.3,0) -- (10.9,0);
  \node[note, anchor=west] at (11.05,0) {wide WAN link};
  \node[fpga] at (13.7,0) {};
  \node[note, anchor=west] at (13.95,0) {edge FPGA};
  \draw[tie] (16.1,0) -- (16.7,0);
  \node[note, anchor=west] at (16.85,0) {clique tie};
\end{scope}
\end{tikzpicture}}
\caption{an illustration of the paper's proposal, over seven islands of compute on an irregular wide area network (WAN). Each island is connected to the network through bandwidth-light access links, but the core mesh is supported by bandwidth-heavy links. In each round, the islands are partitioned into disjoint cliques to synchronize model parameters, e.g. in the first round the islands are partitioned into triangles and in the second round the islands are partitioned into pairs and a leftover triangle. Within a clique in a given round, each sender emits a single stream which is replicated by multicast to all destinations. Each receiver receives its streams through an FPGA that aggregates the information before transmitting it over the access link, so that the final hop carries a single merged stream regardless of number of original senders.}
\label{fig:rotating}
\end{figure}

In principle, these two components are standalone. For instance, one could deploy multicast and FPGAs across a WAN but retain standard synchronization schedules. Alternatively, one could use the synchronization algorithm over a network that lacks such technologies. But this would, in our eyes, waste the potential of each. Better technologies will deliver unimpressive gains if used suboptimally; and the algorithm is explicitly built to leverage multicast and FPGAs when present. Our simulations illustrate how the optimal schedules change as the network's capabilities change.

Indeed, we believe the two components are novel on their own, but again they are more powerful in tandem. Multicast and FPGAs have been used for training within data centers but (to our knowledge) not across a WAN; and there are few synchronization frameworks that are so deeply embedded in the network's properties and capabilities. But together, they offer a complete and self-contained proposal for training the next generation of models.

The paper proceeds as follows. Section \ref{sec:related} discusses related literature and positions our contribution. Section \ref{sec:network} presents the systems proposal, i.e. usage of multicast and FPGAs in a WAN. Section \ref{sec:algorithm} describes the algorithms proposal, i.e. how to construct the optimal synchronization schedule to maximize information flow. Section \ref{sec:simulation} demonstrates both, in a setting inspired by the DoubleZero network (a live programmable WAN equipped with both technologies). Section \ref{sec:conclusion} concludes.

\section{Related Work}
\label{sec:related}
All modern models are trained across parallel GPUs, making the aggregation of local gradients a first-order concern. Early systems used centralized parameter servers \citep{dean2012large,li2014parameterserver}, but those became bottlenecks in turn; and so many modern approaches have turned to decentralized collectives like the ring all-reduce of \citet{patarasuk2009ring}. Over time and within a single data center, a variety of techniques have further optimized that aggregation step, including having the network itself do the work. SHArP \citep{graham2016sharp} and SwitchML \citep{sapio2021switchml} are two of the best-known, where aggregation is done in the switch. Critically, these require tight and synchronous timing, as switches have almost no memory or flexibility.

Given these limitations, \citet{gebara2021panama} introduced PANAMA, which moved aggregation within data centers onto FPGAs that then distributed aggregated updates over multicast. In many ways, this is the systems inspiration for our paper. Critically, though, PANAMA was designed for within-data center usage, where workers were colocated and so enjoyed negligible latency and plentiful bandwidth. Our paper extends it to the WAN setting, where none of these conditions hold.\footnote{There are many other relevant papers in this domain beyond these landmark ones. For instance, to address multicast reliability gaps, \citet{khalilov2024broadcast} implement recovery logic in NICs. ATP \citep{lao2021atp} allows switches to serve multiple tenants in parallel. NetReduce \citep{liu2023netreduce} similarly moves aggregation from a switch into an FPGA without terminating the RDMA transport. TACCL \citep{shah2023taccl} synthesizes compilers for custom aggregation algorithms, tailored to specific topologies.}

Once distributed training moves from a data center to a wide area network, four key challenges emerge. First, bandwidth limitations are severe. A frontier model with a trillion one-byte parameters would naively require links on the order of ten terabits per second to exchange full updates each second. Second, communication latencies over such topologies grow from microseconds to milliseconds, ensuring that islands are always operating with stale parameters (or idling their hardware while waiting). Third, WAN topologies are highly heterogeneous, with asymmetric bandwidths and latencies between islands. This bedevils any scheme that assumes concurrent data arrivals. Fourth, although largely out of scope for our paper, reliability becomes more salient, with nodes sometimes churning and links sometimes failing.\footnote{\citet{keiblinger2025pccl} address fault tolerance head-on, with a collective communications library to tolerate workers joining, leaving, and failing during a training run.} These challenges prevent a naive application of the within-data center techniques to the across-data center setting.

The general response of the field has been to focus on the algorithms over the hardware, across three margins. The first margin is to communicate less often, e.g. federated averaging \citep{mcmahan2017fedavg} or its modern successors DiLoCo \citep{douillard2023diloco} and Streaming DiLoCo \citep{douillard2025streaming}. These assume poorly-connected islands that deliberately take multiple training steps before sending parameter updates around. The second margin is to communicate less volume, i.e. shrinking messages themselves through quantization as in \citet{seide2014onebit} or \citet{alistarh2017qsgd}, sparsification as in \citet{lin2018dgc}, and low-rank projection as in \citet{vogels2019powersgd}. The third margin is to communicate partially.

This third margin of partial communication, where communication happens regularly but in limited and imperfect ways, is where our paper's algorithmic contribution lives. This margin descends from the gossip literature \citep{lian2017dpsgd}, where workers average with a subset of peers only. Our paper is inspired by two particular contributions here. First, \citet{wang2019matcha} introduced MATCHA, which decomposes a graph into matchings (i.e. disjoint pairs of workers) and activates them optimally given a communication budget. Second, \citet{kim2025halos} designed HALoS as a hierarchical system that is geographically-aware and asynchronous, wherein workers interact with a nearby parameter server and those servers interact over longer distances.\footnote{Other key contributions in this domain include lock-free training \citep{niu2011hogwild}, asynchronous DiLoCo \citep{liu2024asynclocal}, and one-peer exponential graphs \citep{ying2021exponential}. There are two especially close to our work: \citet{shen2025bandwidthaware} optimize a mixing topology under static network conditions, and \citet{koneputugodage2026factored} create a hybrid algorithm of gossip and DiLoCo.} Our paper extends both in multiple ways: multi-party cliques rather than two-party matchings, non-hierarchical exchanges that remain geographically aware, and an endogenous communication budget that comes from the network's own topology rather than given as an external constraint.

\section{Network Technology}
\label{sec:network}
Model training within a data center enjoys rich connectivity, as modern clusters connect workers with one another with bandwidth of 400+ Gbps and latency of only a few microseconds. But training across a wide area network enjoys none of that, and instead suffers limited bandwidth, asymmetric connectivity, and tens or hundreds of milliseconds of latency.

However, network technologies can help relieve some bandwidth constraints. The primary concern is the ingress and egress links to the network, i.e. the links that connect islands of compute to the general WAN. These are typically very expensive, as they must navigate urban environments and are hotly contested. Many cloud providers build their entire pricing model around these links, offering cheap compute and storage but expensive egress (and sometimes ingress) to the facility. Thus, the most impactful systems reduce pressure here, by having the network replicate and aggregate information itself rather than having the islands do that directly.

\begin{figure}[htbp]
\centering
\resizebox{0.9\textwidth}{!}{%
\begin{tikzpicture}[
  >=Stealth,
  worker/.style={draw, rounded corners=3pt, fill=white, minimum width=2.05cm,
                 minimum height=0.8cm, font=\scriptsize, inner sep=2pt},
  dev/.style={draw, diamond, aspect=1, fill=gray!25, minimum width=1.3cm,
              minimum height=1.3cm, inner sep=0pt},
  flow/.style={->, line width=0.5pt, black},
  rflow/.style={->, >={Stealth[length=3.2mm,width=2.6mm]}, line width=0.9pt,
                dash pattern=on 3.2pt off 2.2pt, black},
  lbl/.style={font=\tiny, fill=white, inner sep=1pt, midway}
]
\node[worker] (wa)  at (2.16,7.74)  {Island A};
\node[dev]    (hub) at (4.05,3.78)  {};
\node[dev]    (db)  at (8.28,6.48)  {};
\node[dev]    (dc)  at (8.28,1.08)  {};
\node[worker] (wb)  at (11.09,7.74) {Island B};
\node[worker] (wc)  at (11.09,-0.18){Island C};
\node[dev]    (dd)  at (4.05,-0.35) {};
\node[worker] (wd)  at (4.05,-2.75) {Island D};

\draw[flow] (wa.south east) to[bend left=15]  node[lbl]{model A}    (hub.88);
\draw[flow] (hub.110)       to[bend left=15]  node[lbl]{full model} (wa.south);

\draw[flow]  (hub.66) to[bend left=24] node[lbl]{model A} (db.192);
\draw[flow]  (hub.46) to[bend left=6]  node[lbl]{model D} (db.213);
\draw[flow] (db.230) to[bend left=17] node[lbl]{model B} (hub.24);

\draw[flow]  (hub.-66) to[bend right=24] node[lbl]{model A} (dc.168);
\draw[flow]  (hub.-46) to[bend right=6]  node[lbl]{model D} (dc.147);
\draw[flow] (dc.130)  to[bend right=17] node[lbl]{model C} (hub.-24);

\draw[flow] (db.248) to[bend right=17] node[lbl]{model B} (dc.112);
\draw[flow] (dc.68)  to[bend right=17] node[lbl]{model C} (db.292);

\draw[flow]  (hub.240) to[bend right=28] node[lbl, pos=0.28]{model A} (dd.120);
\draw[flow]  (hub.262) to[bend right=0]  node[lbl, pos=0.52]{model B} (dd.98);
\draw[flow]  (hub.284) to[bend left=28]  node[lbl, pos=0.76]{model C} (dd.76);
\draw[flow] (dd.150)  to[bend left=48]  node[lbl, pos=0.55]{model D} (hub.212);

\draw[flow] (dd.244) to[bend right=16] node[lbl]{full model} (wd.150);
\draw[flow] (wd.30)  to[bend right=16] node[lbl]{model D}    (dd.-64);

\draw[flow] (db.105)        to[out=100,in=170] node[lbl,pos=0.62]{full model} (wb.west);
\draw[flow] (wb.south west) to[out=250,in=25]  node[lbl,pos=0.55]{model B} (db.25);

\draw[flow] (dc.255)        to[out=260,in=190] node[lbl,pos=0.62]{full model} (wc.west);
\draw[flow] (wc.north west) to[out=110,in=335] node[lbl,pos=0.55]{model C} (dc.335);
\end{tikzpicture}}
\caption{an illustration of the network technologies over four islands of compute. Each island emits its model once, and network devices replicate streams as they proceed to other islands. Each island's device has an FPGA that aggregates inbound streams to deliver a single full model over the final hop.}
\label{fig:overview}
\end{figure}
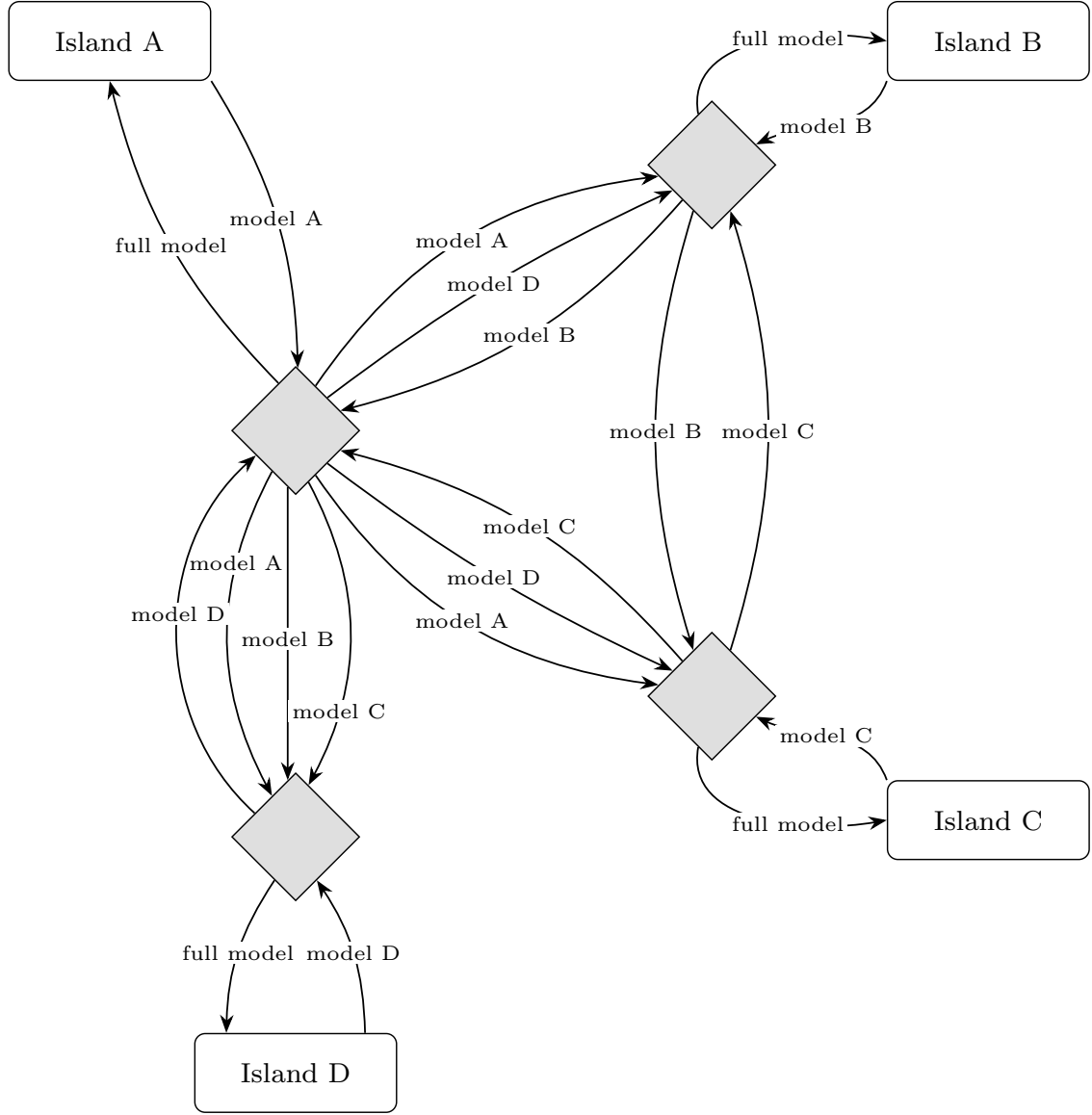

There are two key technologies. On the outbound side, multicast is the key technology that allows an island to send a single payload; and the network replicates it as needed en route to receivers. This preserves bandwidth both across the egress links and the core of the network. On the inbound side, FPGAs aggregate inbound streams at the edges of the network, which allows the network to send a single payload to a receiver over the ingress link. Figure \ref{fig:overview} illustrates the design. These are both well-established within data centers \citep{sapio2021switchml, gebara2021panama} and so the innovation is implementing them to support training over the WAN.
 
\subsection{Multicast}
In a training system with $N$ islands, data must travel from any given island to all others. The default mode of travel is unicast, i.e. the sender sends a stream to each individual receiver. This means that the sender must send $N-1$ copies of the data, and that volume imposes pressure on the egress link. There are likely many similar redundancies within the core of the network, as duplicate streams travel the same links to receivers that are situated close to one another.

\begin{figure}[htbp]
\centering
\begin{tikzpicture}[
  islbox/.style={rectangle, rounded corners=4pt, draw=black, fill=white,
                 minimum height=7mm, inner sep=5pt, font=\scriptsize},
  rtr/.style={circle, draw=black, fill=gray!40, minimum size=13pt, inner sep=0pt},
  st/.style={-{Stealth[length=2.2mm]}, line width=0.7pt, draw=black},
  note/.style={font=\scriptsize, align=center},
  mlab/.style={font=\tiny},
  ptr/.style={draw=gray!70, dashed, line width=0.4pt}
]
\begin{scope}
  \node[note, font=\bfseries\small, anchor=west] at (-1.1,2.5) {(a) unicast};
  \node[islbox] (S)  at (0,0) {Island 1};
  \node[rtr]    (P)  at (2.2,0) {};
  \node[islbox] (R1) at (5.4,1.7) {Island 2};
  \node[rtr]    (Q)  at (3.9,-1.1) {};
  \node[islbox] (R2) at (5.4,-0.6) {Island 3};
  \node[islbox] (R3) at (5.4,-2.0) {Island 4};
  \draw[st] ($(S.east)+(0,0.16)$)  to[bend left=16]  ($(P.west)+(-0.02,0.12)$);
  \draw[st] (S.east)               --                 (P.west);
  \draw[st] ($(S.east)+(0,-0.16)$) to[bend right=16] ($(P.west)+(-0.02,-0.12)$);
  \node[mlab, anchor=south] at (1.1,0.42) {3 copies};
  \draw[st] (P) -- ($(R1.west)+(0,-0.05)$);
  \draw[st] ($(P.south east)+(0,0.06)$) to[bend left=12]  ($(Q.north west)+(0,0.10)$);
  \draw[st] ($(P.south east)+(-0.06,-0.04)$) to[bend right=12] ($(Q.west)+(-0.02,-0.04)$);
  \node[mlab, anchor=north east] at (2.85,-0.85) {2 copies};
  \draw[st] (Q) -- ($(R2.west)+(0,-0.05)$);
  \draw[st] (Q) -- ($(R3.west)+(0,0.05)$);
\end{scope}
\begin{scope}[xshift=8.1cm]
  \node[note, font=\bfseries\small, anchor=west] at (-1.1,2.5) {(b) multicast};
  \node[islbox] (S)  at (0,0) {Island 1};
  \node[rtr]    (P)  at (2.2,0) {};
  \node[islbox] (R1) at (5.4,1.7) {Island 2};
  \node[rtr]    (Q)  at (3.9,-1.1) {};
  \node[islbox] (R2) at (5.4,-0.6) {Island 3};
  \node[islbox] (R3) at (5.4,-2.0) {Island 4};
  \draw[st] (S.east) -- node[mlab, above=1pt] {1 copy} (P.west);
  \draw[st] (P) -- ($(R1.west)+(0,-0.05)$);
  \draw[st] (P) -- node[mlab, below=2pt, sloped] {1 copy} ($(Q.north west)+(0.02,0.06)$);
  \draw[st] (Q) -- ($(R2.west)+(0,-0.05)$);
  \draw[st] (Q) -- ($(R3.west)+(0,0.05)$);
  \node[note] (rep1) at (2.1,1.9) {replicated at\\the branch point};
  \draw[ptr] (rep1.south) -- ($(P.north)+(0,0.06)$);
  \draw[ptr] (rep1.east) to[bend left=20] ($(Q.north)+(0.05,0.08)$);
\end{scope}
\end{tikzpicture}
\caption{unicast versus multicast over one small topology, with one island sending to three. Under unicast, the sender transmits a separate copy per recipient and so many links carry duplicate copies. Under multicast, the sender transmits once and the network replicates it wherever the distribution tree branches, so that every link carries exactly one copy.}
\label{fig:multicast}
\end{figure}
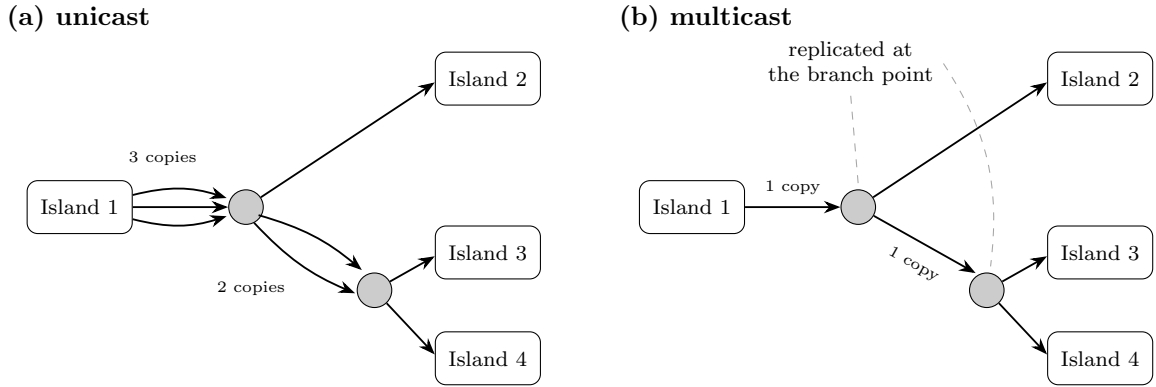

Multicast lets the network, rather than the sender, duplicate data; and in turn the network duplicates data as efficiently as possible. Formally, rather than addressing each recipient individually, a sender sends a stream to a group, and recipient islands subscribe to the groups they wish to receive. The network then builds a distribution tree from the sender to all subscribers, and duplication happens at each branch of the tree \citep{deering1990multicast}. This means that replication happens downstream, relieving pressure on the egress link and eliminating all other forms of duplication. Figure \ref{fig:multicast} visualizes this for a simple four-island topology.

Multicast is a mature technology, but it is only deployed on coordinated networks rather than the public internet. On the public internet, routers belong to independently-administered operators and it would be infeasible to coordinate group membership and distribution trees. Coordinated networks, by contrast, can offer this.

The caveat to multicast is that it precludes using TCP. This means that the transport does not have a completeness guarantee, and so lost packets must be handled through other means. This is often handled by either having the recipient tolerate loss, or using sequence numbers and a side channel for requesting missing sequence IDs. For low bandwidth use cases, a periodic snapshot can be sent so any missing packet is replaced by newer copy shortly. We return to this problem in Section \ref{subsec:packetloss}, as FPGAs can offer some defenses against this issue too.

\subsection{FPGAs}
FPGAs complement multicast, by reducing bandwidth consumption from the WAN to the island. Specifically, in an $N$ island setup, each island would ordinarily receive $N-1$ streams from other islands, and these streams -- which are distinct -- would ordinarily put pressure on the island's ingress link. FPGAs allow for in-network aggregation before reaching the ingress link, so that only a single stream travels that final journey.

This idea is proven within data centers, e.g. SwitchML performs the aggregation within a switch \citep{sapio2021switchml} and other papers move that into an FPGA \citep{liu2023netreduce, gebara2021panama}. But in all of these settings, workers are symmetrically and closely spaced and so streams arrive at concurrent times. This means that their hardware neither contains nor requires meaningful memory (beyond small inbuilt amounts) that can be accessed at line rate. By contrast, islands are differentially spaced over a WAN, rendering any naive implementation unworkable.

Newer FPGAs can relax this constraint, as they incorporate meaningful amounts of high-bandwidth memory. For instance, the AMD Alveo V80 (which is being installed across the DoubleZero network) carries 32 GB \citep{amd2024alveov80}. This allows for aggregation to proceed on streams that arrive at different times. (There are still limits on what this memory can buy, and Section \ref{sec:algorithm} treats the memory as a key input to the right topology.)

Memory must be further managed with careful storage and eviction discipline, so that imperfections and errors do not stall the pipeline. More precisely, each in-flight weight occupies a slot in HBM, keyed by a hashed weight identifier (WID). Each slot carries both the running accumulator and the total count of contributions. The FPGA holds and accumulates into a slot until one of three eviction criteria is met, at which point the slot's contents are forwarded to the island.
\begin{enumerate}
\item Completion: the last island's contribution arrives. This is the expected case, and the aggregated weight is forwarded immediately.
\item Timeout: too much time has passed. This prevents a lagging or stalled island from blocking all the others.
\item Conflict: a new weight arrives whose WID hashes to an occupied slot, and the slot must be surrendered.
\end{enumerate}

The third criterion is inherent to any finite memory, but two factors make it benign. First, by coordinating the hashing scheme with the order in which islands send their weights, the probability of conflict can be driven close to zero. Second, even if a conflict forces an eviction mid-accumulation, the FPGA simply forwards two partial sums per affected weight rather than a single complete one. Because aggregation is associative, the receiving island can combine them and recover the correct result.

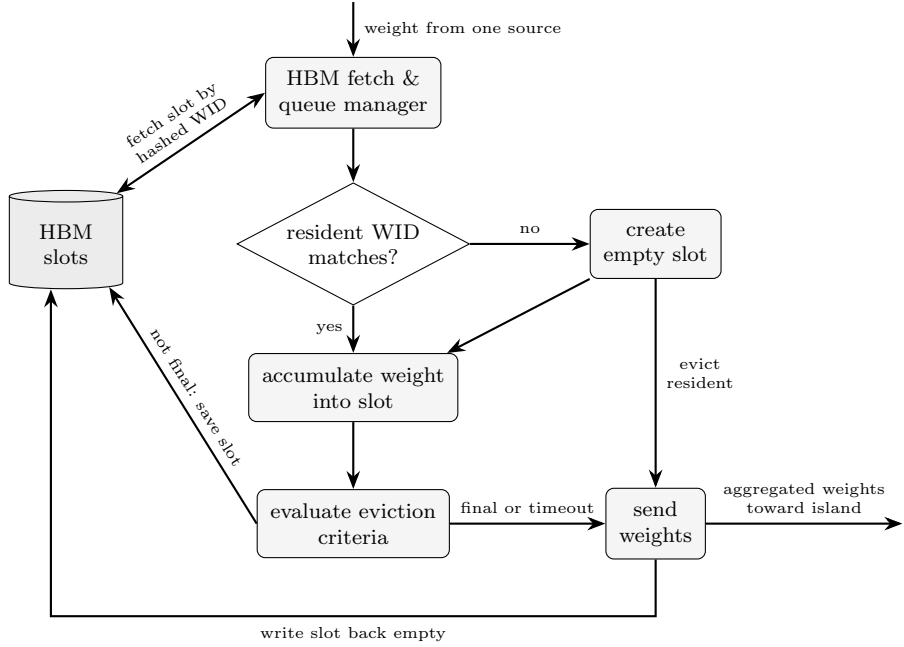
\begin{figure}[htbp]
\centering
\begin{tikzpicture}[
  proc/.style={rectangle, rounded corners=3pt, draw=black, fill=gray!8,
               minimum height=9mm, inner sep=5pt, font=\scriptsize, align=center},
  dec/.style={diamond, aspect=1.9, draw=black, fill=white, inner sep=1.5pt,
              font=\scriptsize, align=center},
  mem/.style={cylinder, shape border rotate=90, aspect=0.18, draw=black,
              fill=gray!15, minimum height=13mm, minimum width=15mm,
              inner sep=2pt, font=\scriptsize, align=center},
  lab/.style={font=\tiny, align=center},
  arr/.style={-{Stealth[length=2.4mm]}, line width=0.8pt},
  barr/.style={{Stealth[length=2.4mm]}-{Stealth[length=2.4mm]}, line width=0.8pt}
]
  \coordinate (in) at (4.6, 9.4);
  \node[proc] (qm)  at (4.6, 8.2) {HBM fetch \&\\queue manager};
  \node[mem]  (hbm) at (0.8, 6.2) {HBM\\slots};
  \node[dec]  (wid) at (4.6, 6.2) {resident WID\\matches?};
  \node[proc] (ce)  at (8.6, 6.2) {create\\empty slot};
  \node[proc] (acc) at (4.6, 4.3) {accumulate weight\\into slot};
  \node[proc] (ev)  at (4.6, 2.5) {evaluate eviction\\criteria};
  \node[proc] (sw)  at (8.6, 2.5) {send\\weights};
  \draw[arr] (in) -- node[lab, right] {weight from one source} (qm);
  \draw[barr] (qm.west) -- node[lab, above, sloped] {fetch slot by\\hashed WID} (hbm.north east);
  \draw[arr] (qm) -- (wid);
  \draw[arr] (wid) -- node[lab, left] {yes} (acc);
  \draw[arr] (wid) -- node[lab, above] {no} (ce);
  \draw[arr] (ce) -- node[lab, right, pos=0.45] {evict\\resident} (sw);
  \draw[arr] (ce.south west) -- ($(acc.north east)+(-0.15,0)$);
  \draw[arr] (acc) -- (ev);
  \draw[arr] (ev.west) -- node[lab, above, sloped] {not final: save slot} (hbm.south east);
  \draw[arr] (ev) -- node[lab, above] {final or timeout} (sw);
  \draw[arr] (sw.south) -- ++(0,-0.75) -| node[lab, below, pos=0.25] {write slot back empty}
      ($(hbm.south)+(-0.20,0)$);
  \draw[arr] (sw.east) -- node[lab, above] {aggregated weights\\toward island} ++(2.6,0);
\end{tikzpicture}
\caption{a simplified diagram of the FPGA. Each arriving weight is matched by hashed weight identifier (WID) to a slot in HBM, accumulated, and saved back until an eviction criterion is met. At that point, the aggregated weight is forwarded to the receiving island and the slot is surrendered.}
\label{fig:accumulator}
\end{figure}

Figure \ref{fig:accumulator} presents a simplified diagram of this system. Note that the queue manager requires further complexity, as it tracks the slots in the accumulation pipeline and which HBM fetches are outstanding, so that no incoming weight is ever summed against a stale copy of its slot.
 
\subsection{Accumulator Functions}
\label{subsec:accumulators}
The FPGA accumulates incoming weights, and thus far we have assumed a simple averaging formula. However, any algorithm can be deployed as long as two criteria are met: operations must be commutative and associative, and operations must not require meaningful amounts of state from a previous round. This means that weighted averages (which might require state per island, but not state per weight) are also admissible.

Momentum, which DiLoCo \citep{douillard2023diloco} utilizes, does not naively fit in this framework, as it would require state per weight. The simplest workaround is to have islands propagate weights and their states when exchanging information, but this doubles network bandwidth utilization and similarly halves the effective memory in the FPGA. However, momentum can be computed at the sender level (applied to an island's own stream before transmission, which works because the momentum recurrence is linear with a constant coefficient and so commutes with averaging) or at the receiver level (applied to the aggregated weights after arrival, which is where DiLoCo's outer optimizer applies it), rather than in-flight over the network. In other words, momentum can still be utilized by the system, but just outside the scope of the network.

\subsection{Packet Loss}
\label{subsec:packetloss}
One wrinkle in the real world is packet loss, even if such losses are small in a private WAN. Typical training systems handle this with TCP, but TCP is incompatible with both multicast and in-network aggregation. As such, this system must be designed to tolerate loss without impairing convergence or stalling the pipeline.\footnote{\citet{khalilov2024broadcast} demonstrate an example of using hardware to implement recovery logic within data centers.}

The key observation is that loss is asymmetric in its severity. A packet lost between a sending island and the receiving island's FPGA is a single island's contribution to a handful of weights, and most forms of distributed training are robust to occasional gaps of this kind. A packet lost between the receiving island's FPGA and the island itself is much more costly, as it contains the combined contribution of every island in the clique. We therefore modify the FPGA design to protect the post-aggregation hop, which is both more important and easier to protect.

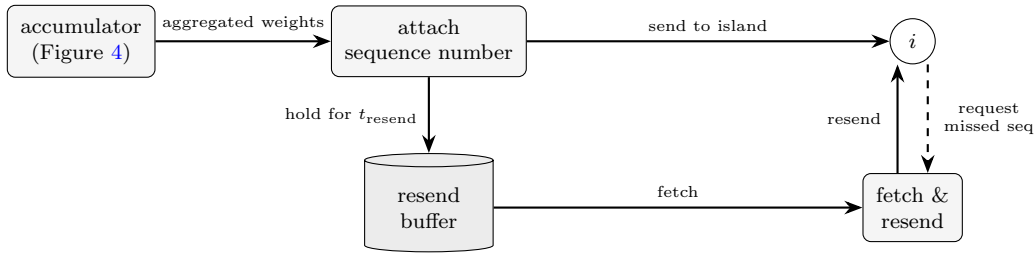
\begin{figure}[htbp]
\centering
\begin{tikzpicture}[
  proc/.style={rectangle, rounded corners=3pt, draw=black, fill=gray!8,
               minimum height=9mm, inner sep=5pt, font=\scriptsize, align=center},
  mem/.style={cylinder, shape border rotate=90, aspect=0.18, draw=black,
              fill=gray!15, minimum height=13mm, minimum width=17mm,
              inner sep=2pt, font=\scriptsize, align=center},
  isl/.style={circle, draw=black, minimum size=17pt, inner sep=1pt, font=\scriptsize},
  lab/.style={font=\tiny, align=center},
  arr/.style={-{Stealth[length=2.4mm]}, line width=0.8pt},
  nack/.style={arr, dashed}
]
  \node[proc] (agg) at (0.6, 2.4) {accumulator\\(Figure \ref{fig:accumulator})};
  \node[proc] (seq) at (5.2, 2.4) {attach\\sequence number};
  \node[isl]  (dst) at (11.6, 2.4) {$i$};
  \node[mem]  (buf) at (5.2, 0.2) {resend\\buffer};
  \node[proc] (fr)  at (11.6, 0.2) {fetch \&\\resend};
  \draw[arr] (agg) -- node[lab, above] {aggregated weights} (seq);
  \draw[arr] (seq) -- node[lab, above] {send to island} (dst);
  \draw[arr] (seq) -- node[lab, left] {hold for $t_{\text{resend}}$} (buf);
  \draw[arr] (buf.east) -- node[lab, above] {fetch} (fr.west);
  \draw[nack] ($(dst.south)+(0.20,0)$) -- node[lab, right=2pt] {request\\missed seq} ($(fr.north)+(0.20,0)$);
  \draw[arr] ($(fr.north)+(-0.20,0)$) -- node[lab, left=2pt] {resend} ($(dst.south)+(-0.20,0)$);
\end{tikzpicture}
\caption{loss recovery on the post-aggregation hop. The FPGA numbers each outbound packet and retains a copy until the recipient could have requested it.}
\label{fig:resend}
\end{figure}

Specifically, we propose that the FPGA attach a sequence number to each outbound packet and retain a copy in a resend buffer for a window $t_{\text{resend}}$. This window must be at least the round trip to the recipient island, plus some processing time; and because the FPGA sits at the network edge, this round trip time is short. If the recipient island detects a gap in sequence numbers, it requests the missing packet, which the FPGA fetches and resends. Figure \ref{fig:resend} illustrates.
 
In principle, loss between a \textit{sending} island and FPGA can be handled by the same system, but the benefits are smaller relative to the costs. Each lost packet here only carries a single island's contribution, which is less important. Moreover, the round trip between the sending island and the FPGA can potentially be much longer, which forces a longer holding window. In turn, the FPGA must either hold slots open for longer, which consumes memory more aggressively, or the system must accept that re-sent weights will often find their slots already evicted by conflicting arrivals. We thus do not recommend this except in systems that are heavily over-provisioned in memory.

\section{The Synchronization Schedule}
\label{sec:algorithm}
Section \ref{sec:network} demonstrates how certain network technologies can be brought to the WAN in support of distributed training. But how exactly should they be used? A naive implementation might, for instance, do an all-to-all exchange of information from every island of compute at every step. This may be feasible (and in some cases, optimal); but it may be inefficient or infeasible for topologies with distant or highly asymmetric spacing.

This section structures the problem, by setting up a well-defined objective function and characterizing the optimal exchange of information under it. The solution generally structures these islands into small and rotating cliques of information exchange, much like a dance schedule.

\subsection{Setup}
For simplicity, we initially assume total symmetry in everything but the distances between islands of compute. There are $N$ homogeneous islands of self-contained compute, where each holds a full copy of the model and an equal share of the training data. The islands each run through $H$ local steps, where each step takes $\tau$ seconds, and then synchronize (while concurrently continuing to train for the next round). This defines $T$ as the round size in seconds, i.e. $T = H \tau$, and $T$ is indeed the key decision variable in this framework.

Each synchronization round is indexed by $k = 1, \dots, K$. During the synchronization process, the islands are partitioned into cliques; and each clique's members share their parameter states with one another.\footnote{Cliques can be considered the multi-party generalizations of MATCHA's pairwise matchings \citep{wang2019matcha}.} These are averaged, such that each member of a clique starts the next training round with the common merged state to add to its continuously-progressing local state.

Note that while islands share parameter states, the bookkeeping in this section is with respect to \textit{changes} in parameters, created by a given island and born during a particular round. We also refer to these as generations. The physical relay of states is important for propagation via intermediaries, e.g. islands can synchronize parameters without synchronizing directly. But the accounting in changes is important for reasoning through notions of freshness and staleness, as changes have definitive birthdays while states do not. The two representations are linked because averaging is linear, so any island's state is identically a weighted portfolio of every generation.

\subsection{Synchronization Schedules}
The synchronization schedule for round $k$ is described as an $N \times N$ mixing matrix $W(k)$ where each row is the mixture from which island $i$'s new state is constructed. For a partition of cliques and given this simplified case, $W(k)$ is block-structured, with a uniform block per clique and zeroes everywhere else. For instance, suppose in a three-island setup, at round $k$, the first two islands form a clique and the third is its own single-member clique. This would yield the following matrix.
\begin{equation}
W(k) = \begin{bmatrix} \frac{1}{2} & \frac{1}{2} & 0 \\ \frac{1}{2} & \frac{1}{2} & 0 \\ 0 & 0 & 1 \\ \end{bmatrix}
\label{eq:blockmatrix}
\end{equation}

Note that strictly speaking, nothing in Equation \eqref{eq:blockmatrix} requires such uniformity: any non-negative and symmetric matrix with rows and columns summing to one, i.e. doubly-stochastic, preserves the logic in this section. We do not consider this generalization for now, to keep the exposition simple.

Under Equation \eqref{eq:blockmatrix}, $W(k)$ is a Markov transition matrix. In Equation \eqref{eq:transitionmatrix}, this allows us to trace the evolution of a generation $s$ after $h$ rounds of synchronization, as the $\Phi$ matrix, where the $(ij)$ element tells us the weight at which island $i$ currently holds the parameters from island $j$, born at $s$. This incorporates both direct transfers of information -- where $j$ and $i$ synchronize -- and indirect transfers -- where $j$ synchronizes with $m$, and later $m$ synchronizes with $i$.
\begin{equation}
\Phi_s(h) = W(s+h) \cdots W(s+2) W(s+1)
\label{eq:transitionmatrix}
\end{equation}

\citet{koloskova2020unified} provide the convergence theory for such sequences of mixing matrices. That is, for symmetric and doubly-stochastic $W(k)$, decentralized training converges as long as the products of consecutive mixing matrices contract the dispersion of island states around their mean (i.e. their ``consensus distance''). The survival curves discussed next are precisely how we quantify the speed of this contraction; but in all valid schedules (i.e. schedules with a finite objective function value), the dispersion contracts and fulfills the condition.\footnote{More precisely, any schedule in which every pair of islands synchronizes directly or through intermediaries within some period satisfies this. Note that \citet{koloskova2020unified} also require smooth objectives, bounded gradient noise, and bounded cross-island diversity, the last of which is delivered by our setup of identical islands.}

\subsection{Survival Curve}
The ideal for $\Phi_s(h)$ is a matrix where every entry is $1/N$, as this means that all information exchanged during synchronization has diffused perfectly and evenly, as if the islands were co-located.

In practice, of course, there are deviations from this ideal, and there is some survival curve of those deviations. This will form the heart of the objective function, as it is precisely the quantity that the algorithm is trying to minimize. Specifically, we define the error of a given generation $s$ measured after $h$ rounds of synchronization as Equation \eqref{eq:survival}.\footnote{The absolute values per matrix element are justified for two reasons. First, within a single generation, weights across islands always sum to one, so all disagreements are offset in totality. The absolute value recovers the disagreement. Second, across generations, the justification is statistical: with independent data, each round's increment is independent of every other's; and in a high-dimensional parameter space, they are essentially orthogonal. Thus, the absolute value operator again recovers disagreement that might otherwise be silently collapsed.}
\begin{equation}
e_s(h) = \frac{1}{2N - 2} \sum_i \sum_j \left| [\Phi_s(h)]_{ij} - \frac{1}{N} \right|
\label{eq:survival}
\end{equation}

Note that the prefactor in Equation \eqref{eq:survival} is chosen so that $e_s(0) = 1$, i.e. a newborn generation (which is held at weight one by its island, and zero by all others) scores exactly one. From this, $e_s(h)$ reads directly as a fraction of the surviving disagreement after $h$ rounds from birth, for generation $s$. Graphically, this forms a survival curve that starts at one and is driven to zero through mixing.\footnote{As an aside, notice that settlement is absorbing in this formulation. Once a generation reaches $1/N$ weight everywhere, every subsequent average is one of identical numbers. This means the survival curves have finite support, so the area underneath is a finite sum. In practice, schedules that never settle decay geometrically, and so the area underneath the curve can be approximated to any precision.}

The final observation is that, assuming the schedule has a recurring period of $P$, all curves born at the phase $s \bmod P$ share the same decay curve as they experience identical synchronization schedules. Thus, there are $P$ classes of generations and so $P$ distinct survival curves. It will be helpful later to construct the average area across all distinct phases within that period, which we define as $A$ in Equation \eqref{eq:perioda}.

\begin{equation}
A = \frac{1}{P} \sum_{p = 0}^{P - 1} \sum_{h \ge 0} e_p(h)
\label{eq:perioda}
\end{equation}

In words, $A$ is the mean lifetime of a unit of disagreement for a given schedule, and together with $T$ is a key input in the objective function.

\subsection{The Objective Function}
With these primitives established, we are finally ready to define the objective function. The principle behind this objective function is information \textit{staleness}, i.e. an aggregation of the various survival curves of the deviation from the optimal. This is because staleness makes the model training less efficient, and we assume that this loss is proportional to the aggregate staleness via some drift penalty.

To construct the objective function, we first begin with defining the loss as the sum of all survival curves over all generations, each from its birth to the final round. This, in Equation \eqref{eq:sumloss}, simply sums up all disagreement over the course of the training run.
\begin{equation}
\mathcal{L} = \sum_{k=1}^{K-1} \sum_{h = 0}^{K-k-1} e_k(h)
\label{eq:sumloss}
\end{equation}

The inner sum is the generation $k$'s lifetime disagreement, clipped at the end of the training run. Under periodicity, the \textit{unclipped} summation is a function of the birth phase only. Indeed, notice that $A \cdot P$ from Equation \eqref{eq:perioda}, i.e. the sum of area under all survival curves for a period, looks similar to this --- except it has fewer outer terms (unique phases $P$ rather than $K-1$ rounds) and more inner terms $h \ge 0$, rather than bounded at $K - k - 1$. Assume for simplicity then that $P$ divides $K$ exactly; then the loss can be rewritten as Equation \eqref{eq:simplification1}.

\begin{equation}
\mathcal{L} = \frac{K}{P} PA - \sum_{h \ge 0} e_K(h) -  \sum_{k = 1}^{K-1} \sum_{h \ge K - k} e_k(h)
\label{eq:simplification1}
\end{equation}

The second and third terms in Equation \eqref{eq:simplification1} can be collapsed into a remainder term. The second term is one cohort's full area, and does not grow with the size of $K$. The third term is finite under settlement or geometric decay, and so also does not grow with the size of $K$. Thus, as $K$ gets large, the loss function is dominated by the first term alone, as in Equation \eqref{eq:simplification2}.
\begin{equation}
\mathcal{L} = KA + O(1) \approx KA
\label{eq:simplification2}
\end{equation}

Furthermore, the loss function $\mathcal{L}$ grows without bound as $K$ grows large, as it is a full summation. However, we can normalize this term by $K$, which is a given parameter and common to all solutions, such that the solution-dependent quantity is the rate $A$. In words, $A$ is simply the average cohort's lifetime disagreement within a given schedule.

The final step is to restore physical units to Equation \eqref{eq:simplification2}, and convert $A$ from round time into seconds. The key decision variable here is $T$, which is the length of the round, and so multiplies through. In addition, notice that the \textit{creation} of parameter innovations -- which happens before the round of synchronization -- carries its own staleness, for these parameters were updated uniformly over a round of length $T$ and so are $T/2$ old as they enter the synchronization steps.

This allows us to construct the main objective function and constraint as Equation \eqref{eq:objective}.
\begin{equation}
\min_{T, \{W(k)\}} T \left(A+ \frac{1}{2}\right)~~\text{s.t. every $W(k)$ is feasible at $T$}
\label{eq:objective}
\end{equation}

$T$ is the key variable in Equation \eqref{eq:objective}. When it is longer, it directly drives the objective function up; but richer synchronization schedules are available, i.e. the set of feasible $W(k)$ is larger, and so $A$ is reduced. When it is shorter, it directly drives the objective function down; but it only works with simpler schedules, which raises $A$. The objective function here trades off these two.

\subsection{Schedule Feasibility}
The objective function deliberately contains no knowledge of the network. Instead, the network constraints manifest in which schedules $\{W(k)\}$ are feasible, i.e. which synchronization schedules physically fit in $T$. Bad network technologies or topologies force either a larger $T$ or a larger $A$, while good ones achieve improvements on both margins.

To reason about this, we must first define the physical primitives. There are two major constraints: cross-island latency and serialization. Cross-island latencies, defined as $\ell_{ij}$ between islands $i$ and $j$, are of course the travel time over the network between two islands, defined in seconds to  be consistent with other variables. Serialization reflects the fact that the payload cannot enter the network instantaneously. Somewhere along the path of the data (e.g. limitations on the NIC, constraints on the edge connection onto the network, congestion on the cross-network links, etc) there is the tightest bottleneck through which the payload must be metered. This is defined as $D/b$, where $D$ is the payload size (in bytes) and $b$ is the binding pipe's rate (in bytes per second). Crucially, $b$ is both a function of the hardware -- which sets physical limits on capacity -- and the proposed schedule, for a schedule may impose congestion on a link or have insufficient capacity to process an inbound stream of data, and thus tighten the bottleneck further.

The two tools to ease these constraints are multicast and FPGAs. Multicast ensures that no bandwidth is wasted on duplicate streams of the same data, including islands' egress bandwidth onto the network, as replication happens only at branch points as the data travels the network. The FPGAs -- which sit at the network's edge at the islands' ingress points -- aggregate inbound streams, so that the final hop contains a single stream regardless of clique size. (Specifically, they average inbound streams byte position-by-byte position as the streams arrive.) They are endowed with memory $M$ in bytes, which allows for bytes of the same position to arrive at different times.

The synchronization schedule under round time $T$ is comprised of information for each round. First, each round partitions the islands into cliques, i.e. which islands will exchange information. Second, each clique has designated offsets. That is, island $j$ transmits at time $t_j \ge 0$ after its round begins, so that its first byte arrives at island $i$ at $t_j + \ell_{ij}$ and its last byte arrives at $t_j + \ell_{ij} + D/b$. This delay is to help parallel streams arrive at the FPGAs at the same time. Third, each clique is assigned a $b_c$, i.e. the stream rate for that clique, rather than a generic $b$. These are constrained physically by the mechanical limitations imposed by the network capacity, but they can be lowered further if needed to accommodate insufficient memory in the FPGAs. In short, offset delays and stream rates (that are tighter than the network-imposed constraints) are both means to reduce pressure on the memory in FPGAs, by respectively aligning the landing time of bytes and reducing the number of bytes to be processed in tandem.

The schedule $\{W(k)\}$ must then be checked for feasibility. If it meets two conditions, then it is valid, and it can be used to evaluate the objective function.

\begin{enumerate}
\item \textbf{Completion}: the entire exchange of information must be self-contained within its round. That is, the flight of the first byte of data from island $j$ to the arrival of the last at its destination $i$ must complete in $T$, for all islands in the same clique. Mathematically, this can be represented as Equation \eqref{eq:completionconstraint}.
\begin{equation}
(t_j + \ell_{ij}) + D/b_c \le T~~\forall~i, j \in \text{clique $c$},~\forall~c
\label{eq:completionconstraint}
\end{equation}
\item \textbf{Memory}: each FPGA must hold a byte in memory from its arrival of the first stream to its arrival of the last, so that it can perform the aggregation and send it along to its attached island. Mathematically, this can be represented as Equation \eqref{eq:memoryconstraint}.
\begin{equation}
b_c \cdot \left( \max_{j \neq i} (t_j + \ell_{ij}) - \min_{j \neq i} (t_j + \ell_{ij}) \right) \le M~\forall~i \in \text{clique $c$},~\forall~c
\label{eq:memoryconstraint}
\end{equation}
\end{enumerate}

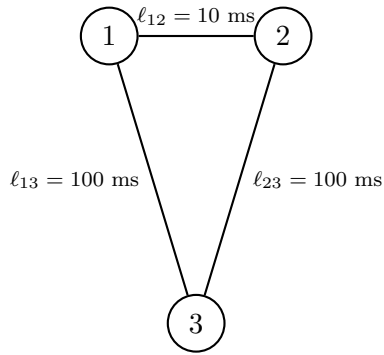
\begin{figure}[!htpb]
\centering
\begin{tikzpicture}[
  island/.style={circle, draw, thick, minimum size=7.5mm, inner sep=0pt},
  lat/.style={font=\scriptsize}
]
\node[island] (i1) at (-1.15,  0)   {$1$};
\node[island] (i2) at ( 1.15,  0)   {$2$};
\node[island] (i3) at ( 0,  -3.8)  {$3$};
\draw[thick] (i1) -- node[lat, above] {$\ell_{12} = 10$ ms} (i2);
\draw[thick] (i1) -- node[lat, left,  xshift=-1pt] {$\ell_{13} = 100$ ms} (i3);
\draw[thick] (i2) -- node[lat, right, xshift= 1pt] {$\ell_{23} = 100$ ms} (i3);
\end{tikzpicture}
\caption{a hypothetical clique, with two islands close to one another and a third far from both. Two extreme solutions exist: all islands send immediately, but the near islands must hold states in memory; or islands stagger sending, which raises completion time but requires no memory.}
\label{fig:triangle}
\end{figure}

Both constraints are in some sense the same, under the hood, for time and memory partially substitute for one another in feasibility. Each constraint formally rations a different resource, but the two constraints are coupled through the shared offsets $t$ and clique stream rate $b$. This can be illustrated using the simple case of a three-island clique, which forms an isosceles triangle in Figure \ref{fig:triangle}.

There are two extreme solutions. The first is for every island to immediately send data at the start of the round, i.e. $t_j = 0~\forall j$. This means that the FPGA at island 1 must hold the fast-arriving stream from island 2 for 90ms in memory whilst the stream from island 3 arrives, which is feasible if either it has sufficient memory or the stream rate for the clique is reduced. The second is for every island to send data at differential times. That is, island 3 begins sending data immediately; but islands 1 and 2 wait for 90ms. In this scenario, each stream arrives at each island at the exact same time, and so no memory is needed and no constraints on the stream rate need be imposed. But this delays the completion of the round by 90ms instead.\footnote{A triangle is a special case where, strictly speaking, no memory is ever needed, as there are three offset variables to solve for three unique pairs of islands. Any larger shape will require some memory. For instance, a square has four offset variables but six unique pairs.}

This more generally illustrates the economics behind feasible schedules. The set of feasible schedules is increasing in round time $T$, as longer rounds absorb longer completion spans, tolerate lower throttled rates, and permit more complex cliques. It is also increasing in the memory $M$, as more memory allows stream rates to rise, offsets to shrink, and again more complex cliques to form. This set grows in jumps at boundary points, which (as the next section discusses) is how the optimal solution is found.

\subsection{Solving the Program}
The objective function in Equation \eqref{eq:objective} seems complex at first glance, especially given all the feasibility constraints on $\{W(k)\}$ in tandem. In practice, it is relatively straightforward to solve.

The key insight is that the solution is always at a boundary point. The objective function is smoothly increasing in $T$ as the outer factor; and decreasing as a step function in $T$ via $A$. In turn, $A$ is only a function of the schedule of cliques. So -- perhaps unsurprisingly -- the optimal will be found at the lowest possible $T$ that can accommodate any particular schedule, as any interior solution only raises $T$ without lowering $A$.

Thus, the algorithm to find the solution iterates over all candidate schedules and mechanically finds the minimum $T$ needed to make them feasible. Then we evaluate the objective function at each, and take the best-performing one.

Specifically, there are five steps to finding the solution. These all assume a fixed and common set of hardware (e.g. memory $M$, link capacities, latencies, etc).

\begin{enumerate}
\item \textbf{Enumerate all candidate schedules.} While there is an infinite number, we should focus on \textit{reasonable} ones, e.g. favoring small and tight cliques first before going to large and complex ones. In the empirical section, we illustrate schedule topologies that seem to perform well, and what features make them especially potent. The following steps then iterate over each schedule individually.
\item \textbf{Construct the stream rate bound per clique.} Each clique's stream rate is bounded by the minimum physical capacity that a network can offer to each data stream within the clique; call this bound $b_c^{\text{max}}$. We first set the bound as the minimum capacity from islands to their local network switches and FPGAs, amongst all islands in a clique. For this particular set of links, there is always a single stream of data traveling in and out, due to multicast technology and FPGA aggregation. Second, congestion in the center of the network can tighten this bound further. We assume that multicast routing always selects the shortest path, and sends exactly one sender-unique stream over each link that it utilizes. (In this setup, the FPGAs are located only on the periphery of the network and so never perform aggregation within the network center.) With these assumptions, we identify the routing for every concurrent clique, and divide up the link's capacity equally amongst all streams using that link. The final per-clique bound is the minimum of the edge capacities and these values.\footnote{There are two observations worth noting. First, each clique's ceiling is a function of the whole concurrent round of cliques rather than just that clique, because of contention between two cliques for the same edge link. Second, a clique can also throttle \textit{itself}. For instance, a clique with multiple senders on one side of a link and multiple receivers on the other side will have to send concurrent streams (one per sender) across that link, and that will decrease capacity for each one.}
\item \textbf{Price each clique's time threshold.} Define $T_c$ as the shortest round at which clique $c$ can be feasible. Despite the seemingly high-dimensional search space, i.e. the actual stream rate and the offsets $t$, this is actually a linear program that can be solved easily. One small substitution is needed, as stream rate $b_c$ enters non-linearly in both constraints but its reciprocal (call it $u_c$) enters linearly. The program is specified in Equation \eqref{eq:cliquelp}.
\begin{equation}
\begin{aligned}
T_c &= \min_{z, \{t\} \ge 0, u_c \ge 1/b_c^{\text{max}}} z &  \\
\text{subject to}~~~& t_j + \ell_{ij} + Du_c \le z & \forall~\text{distinct}~i, j \in \text{clique}\\
& (t_j + \ell_{ij}) - (t_{j'} + \ell_{ij'}) \le Mu_c & \forall~\text{distinct}~i, j, j' \in \text{clique} \\
\end{aligned}
\label{eq:cliquelp}
\end{equation}
The first constraint is Equation \eqref{eq:completionconstraint}; and the second constraint is Equation \eqref{eq:memoryconstraint}. Two limiting cases anchor some intuition. When no memory is available (i.e. $M = 0$), all pairwise landing differences must be zero (which in turn is only feasible for two- or three-island cliques), and the solution $T_c$ is the slowest cross-island latency, serialization latency, and offset latency. When infinite memory is available (i.e. $M \rightarrow \infty$), the second constraint does not bind and so the solution maxes out the clique stream rate and sets the offsets to zero. Jointly, these more generally indicate that $T_c$ starts at the slowest flight in the clique, and -- if memory or bandwidth is insufficient -- it must face extra delays in offsets or serialization.
\item \textbf{Find the minimum $T$ for a schedule.} Since $T$ is the constant round length that must fit all cliques at all times, the maximum $T_c$ value across all cliques for a schedule is the schedule's minimum $T$.
\item \textbf{Score and compare.} The final step is to compute each schedule's $A$ from its survival curves. For a given schedule, this is a purely deterministic calculation in Equation \eqref{eq:perioda} to trace out and aggregate survival curves. Once complete, the objective function in Equation \eqref{eq:objective} can be evaluated for that schedule and the $T$ from the previous step. The smallest objective function across all candidate schedules wins.
\end{enumerate}

Although the best available $A$ moves in steps as $T$ grows, it is interesting to consider what a solution would look like if $A$ was a smooth function of $T$. A direct optimization of Equation \eqref{eq:objective} would note the first order condition sets the elasticity of $A$ with respect to $T$ to $-(1 + 1/(2A))$, which is slightly more than unit elastic. A longer round incurs a fixed cost of propagating more stale data, so the optimal solution must buy a more-than-proportional improvement in mixing to be worthwhile.

\section{System Simulation}
\label{sec:simulation}
We illustrate different schedules in a simulation environment that looks like the DoubleZero network. This demonstrates the value-add of both the systems component and the scheduling component.

\subsection{Setting}
The DoubleZero network is a decentralized physical network associated with the DoubleZero project (\url{https://www.doublezero.xyz}). As of writing, it has fifteen independent contributors connecting over sixty facilities in thirty cities, with links ranging from 10 Gbps of capacity to 100 Gbps. This network employs both multicast technology and spreads FPGAs with high-bandwidth memory across the network. DoubleZero is a project to make the next generation of financial markets, including prediction markets, perpetuals markets, and crypto markets, run at scale and speed.

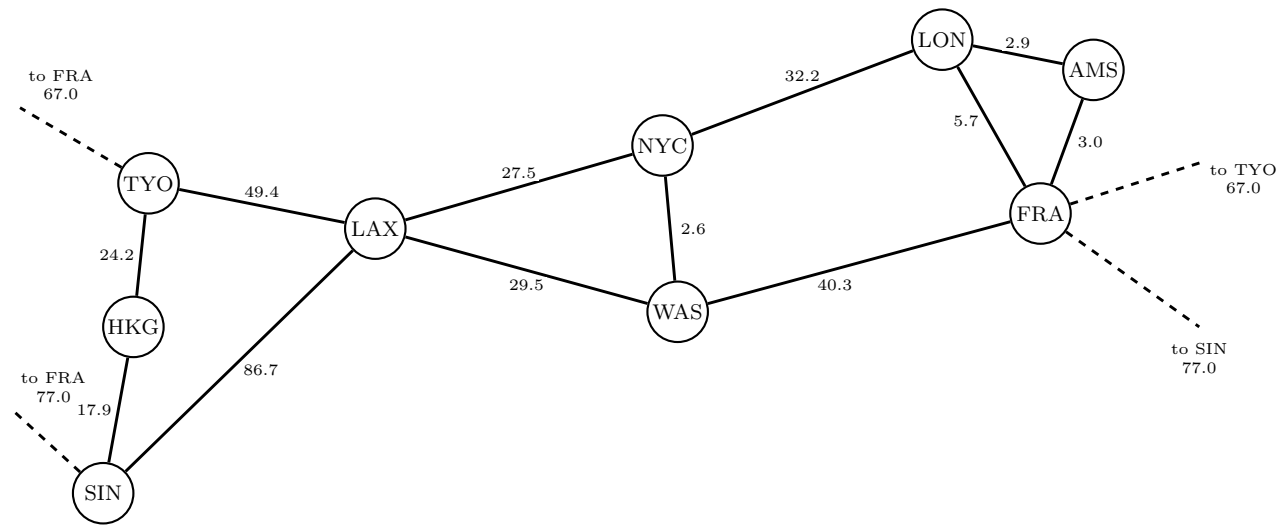
\begin{figure}[htbp]
\centering
\begin{tikzpicture}[
  island/.style={circle, draw, thick, minimum size=8mm, inner sep=0pt,
                 font=\scriptsize},
  lat/.style={font=\tiny, fill=white, inner sep=1pt},
  stublab/.style={font=\tiny, align=center},
  g100/.style={line width=1.1pt},
  g10/.style={line width=1.1pt},
  stub100/.style={line width=1.1pt, dashed},
  stub10/.style={line width=1.1pt, dashed}
]
\node[island] (TYO) at ( 1.9, 5.0) {TYO};
\node[island] (HKG) at ( 1.7, 3.1) {HKG};
\node[island] (SIN) at ( 1.3, 0.9) {SIN};
\node[island] (LAX) at ( 4.9, 4.4) {LAX};
\node[island] (NYC) at ( 8.7, 5.5) {NYC};
\node[island] (WAS) at ( 8.9, 3.3) {WAS};
\node[island] (LON) at (12.4, 6.9) {LON};
\node[island] (AMS) at (14.4, 6.5) {AMS};
\node[island] (FRA) at (13.7, 4.6) {FRA};
 
\draw[g100] (TYO) -- node[lat, above=2pt] {49.4} (LAX);
\draw[g10]  (TYO) -- node[lat, left=1pt]  {24.2} (HKG);
\draw[g10]  (HKG) -- node[lat, left=1pt]  {17.9} (SIN);
\draw[g10]  (SIN) -- node[lat, below right=0pt] {86.7} (LAX);
 
\draw[g100] (LAX) -- node[lat, above=2pt] {27.5} (NYC);
\draw[g100] (LAX) -- node[lat, below=2pt] {29.5} (WAS);
\draw[g100] (NYC) -- node[lat, right=2.5pt] {2.6}  (WAS);
 
\draw[g100] (NYC) -- node[lat, above=3pt] {32.2} (LON);
\draw[g100] (WAS) -- node[lat, pos=0.42, below=2.5pt] {40.3} (FRA);
\draw[g100] (LON) -- node[lat, above=1pt] {2.9}  (AMS);
\draw[g100] (AMS) -- node[lat, right=2.5pt] {3.0}  (FRA);
\draw[g100] (LON) -- node[lat, pos=0.45, left=2pt] {5.7}  (FRA);
 
\draw[stub10]  (TYO) -- ( 0.2, 6.0)
               node[stublab, above right=-1pt] {to FRA\\67.0};
\draw[stub100] (SIN) -- ( 0.1, 2.0)
               node[stublab, above right=-1pt] {to FRA\\77.0};
\draw[stub10]  (FRA) -- (15.9, 5.3)
               node[stublab, below right=-3pt] {to TYO\\67.0};
\draw[stub100] (FRA) -- (15.8, 3.1)
               node[stublab, below=2pt] {to SIN\\77.0};
 
\end{tikzpicture}
\caption{a nine-island topology modeled off the DoubleZero network. Edge labels are one-way latencies in milliseconds; and note that Frankfurt connects to both Tokyo and Singapore over and around Eurasia.}
\label{fig:dz9}
\end{figure}

We model a network that is similar to the DoubleZero network, in Figure \ref{fig:dz9}. This map filters down to nine important cities (three in each region), uses realistic latencies, assumes 100 Gbps of capacity on each link, and assumes 20 Gbps of ingress and egress capacity. We set the model state size to one gigabyte.

\subsection{Illustrative Schedule}
To begin, consider a simple two-phase schedule, called ``rotating triangles.'' In the first phase of this schedule, each region forms a clique: Frankfurt, Amsterdam, and London; New York, Washington, and Los Angeles; and Tokyo, Singapore, and Hong Kong. The second phase of the schedule uses only cross-region cliques, e.g. London, New York, and Singapore; Tokyo, Frankfurt, and Los Angeles; and Amsterdam, Washington, and Hong Kong. Finally, for simplicity, assume that there is no memory capacity at the FPGAs, so parallel data streams must arrive at the same time.

\begin{figure}[htbp]
\centering
\includegraphics[width=1\textwidth]{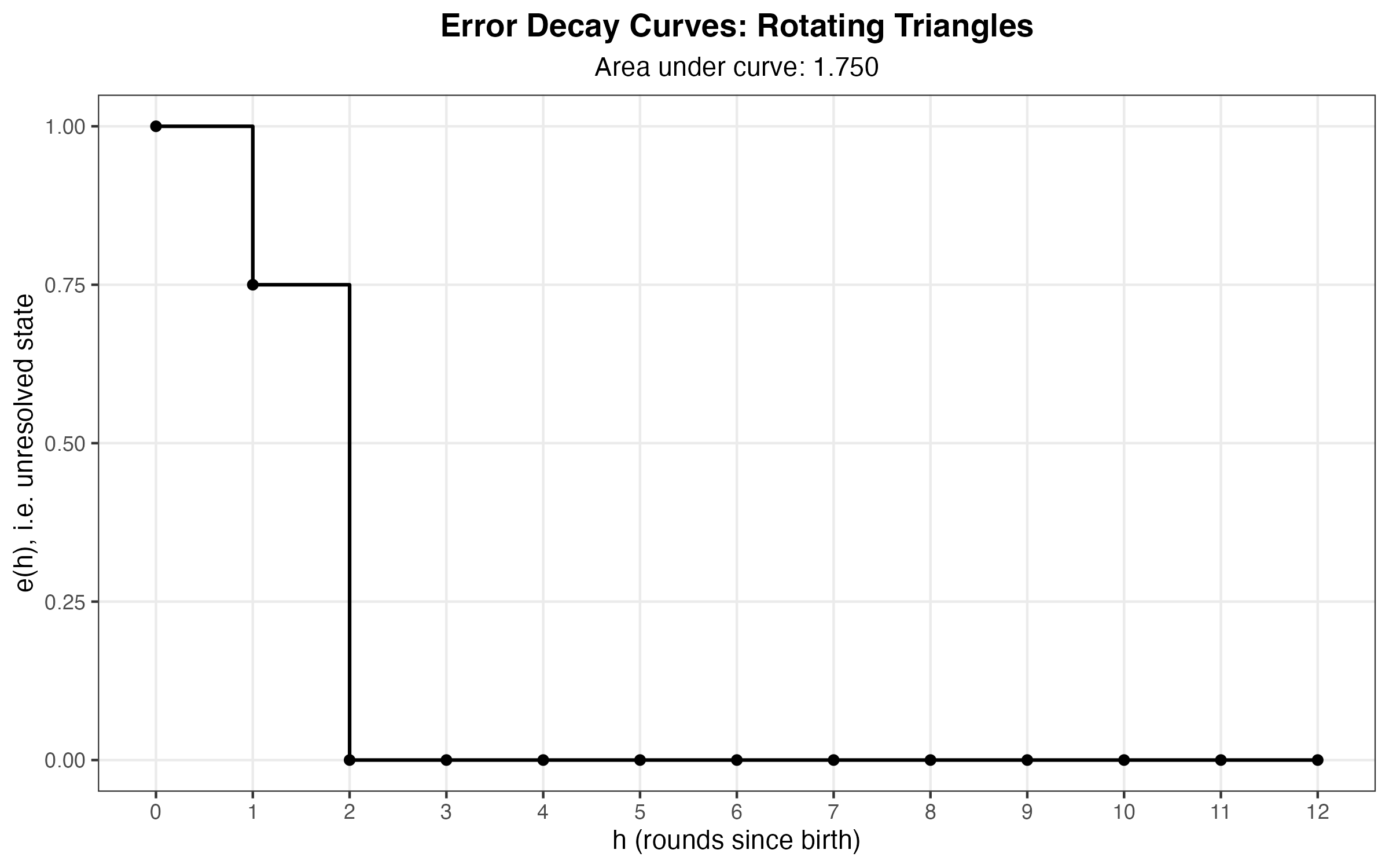}
\caption{the decay of unresolved state through multiple rounds of the triangle schedule.}
\label{fig:area1}
\end{figure}

The objective in Equation \eqref{eq:objective} has two components. The first component is average error per phase, i.e. the $A$ term in Equation \eqref{eq:perioda}. This tells us the quality of mixing, given a certain schedule, without regards to time. (It is measured per cycle, rather than per second.) Figure \ref{fig:area1} demonstrates that this schedule has good mixing properties, with any set of updates being propagated to all islands within two periods, as every island synchronizes either directly or indirectly with every other island. For instance, although New York and Tokyo never synchronize directly, they do so via Los Angeles and Singapore respectively.

The second component is the time needed for this schedule to work. This is computed at 565ms, given the lack of memory and thus need for offsets and delays at the island level. Taken together, this schedule scores 1271 units on the objective function, which is formally in milliseconds-error units.

\subsection{Optimizing Schedules}
For this particular problem, we consider four different schedules.
\begin{enumerate}
\item \textbf{Triangles}: this schedule is described previously.
\item \textbf{Pairs}: this two-round schedule begins by ordering islands in a line, i.e. Singapore, Hong Kong, Tokyo, Los Angeles, Washington, New York, London, Amsterdam, and Frankfurt. It then alternates in pairs: Hong Kong with Singapore in the first iteration, and Hong Kong with Tokyo on the second. (One island is always left out.)
\item \textbf{All-to-All}: this one-round schedule is every island sending data to every other island in every round.
\item \textbf{Regional}: this is a custom schedule that focuses on regional groups, with light mixing between groups. The first round is identical to the round in triangles, where European, North American, and Asian cliques are formed. In the second round, London and New York share a fast path so they form a clique; Frankfurt and Singapore form a clique; and Tokyo and Los Angeles form a clique. Hong Kong, Washington, and Amsterdam are left out.
\end{enumerate}

\begin{figure}[htb]
\centering
\includegraphics[width=1\textwidth]{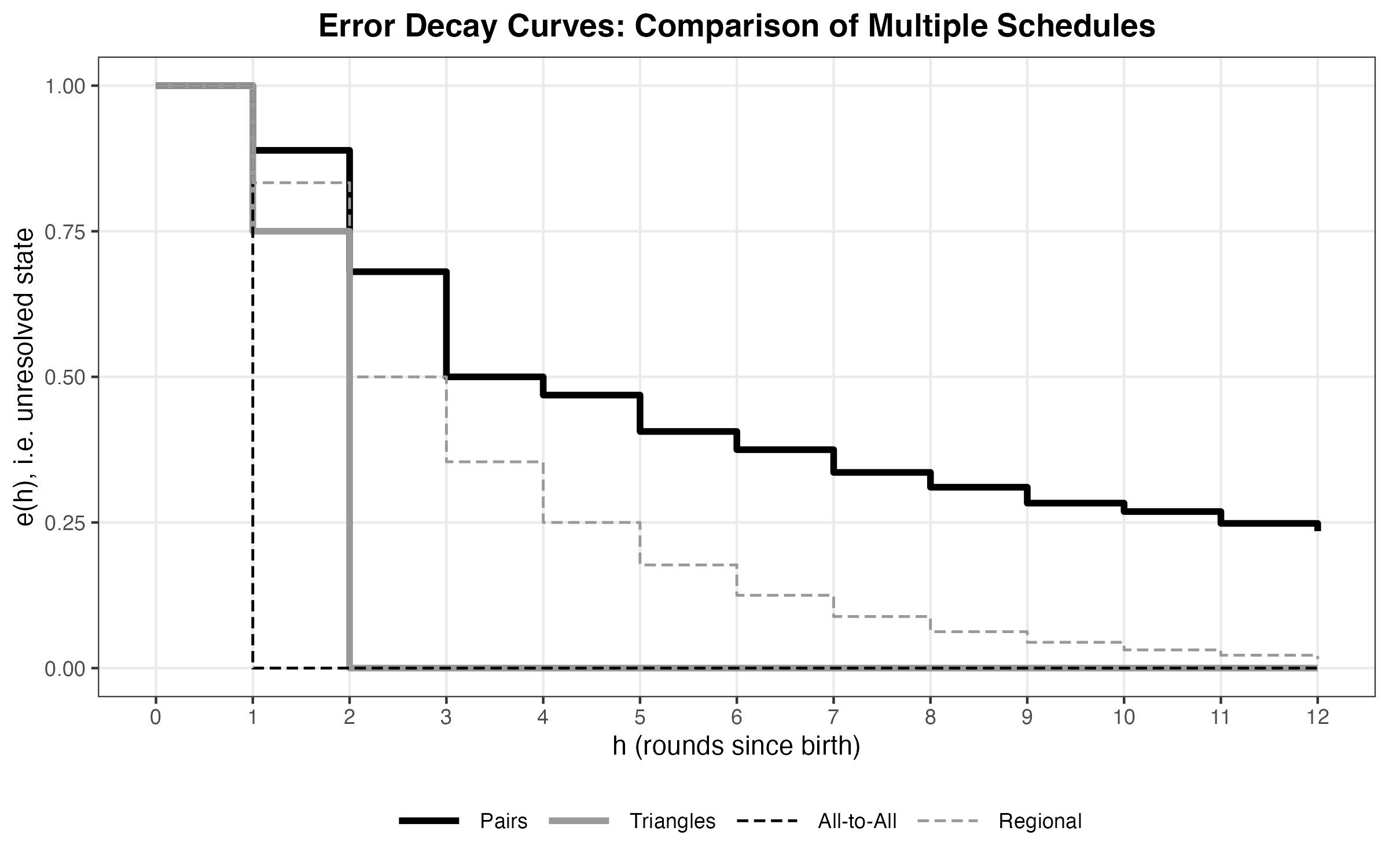}
\caption{the decay of unresolved state through multiple rounds of all four schedules.}
\label{fig:area2}
\end{figure}

Figure \ref{fig:area2} demonstrates the quality of mixing. Unsurprisingly, the all-to-all schedule is perfect on this metric: in each round, every island shares its updates with every other island. The triangle schedule does second best. The regional schedule is third, for it scores highly on regional mixes but it involves slow propagation of those states across continents. Finally, the pair schedule is worst: it takes multiple rounds for one island's updates to reach another, as those updates must flow through multiple neighbors one round at a time.

However, the other component is time. Pairs are slow to propagate updates in round terms, but pairs have fast rounds: only 449ms is needed, which is effectively the latency of the longest link plus the serialization delay. The regional schedule does second-best, at 477ms, for it cuts expensive cross-continental communication down to the minimum; but it still has slightly slow error decay. Triangles does third-best, at 565ms. Finally, the all-to-all schedule is entirely infeasible under this setup, for the FPGAs have no memory and so there is no way to coordinate arrival times between nine senders and nine receivers.

On net, the triangle schedule is best for this setup, for it decays errors quickly at only a bit more latency cost. But the ordering flips for different setups. For instance, if the FPGAs are endowed with sufficient memory (e.g. 32 GB of memory), the all-to-all now becomes entirely feasible with a round time of only 676ms. Although this is still the slowest round time of all schedules, it is not by much; and in tandem, it has phenomenal mixing properties. Thus, under this variant of the baseline schedule, the all-to-all schedule scores best.

\newpage
\subsection{Comparing to Synchronous SGD}
To draw further intuition, we compare two schedules against synchronous stochastic gradient descent \citep{goyal2017accurate}, i.e. the canonical means of distributed training. Under synchronous SGD, each island exchanges updates with every other before proceeding to the next step. In this simulation, we use the same network and endow the FPGAs with 32 GB of memory.

To make the comparison more informative, we look at two variants of synchronous SGD. The first is the canonical one, wherein updates are propagated via a ring all-reduce \citep{patarasuk2009ring} and hosts perform the aggregation. The second is one where synchronous SGD is given access to the network technologies of multicast and FPGAs, for more efficient exchanges. This helps distinguish the value-add of the enhanced hardware from the proposed algorithm. We also look at two variants of our algorithm: the all-to-all schedule, and the rotating triangles schedule.

\begin{figure}[htbp]
\centering
\includegraphics[width=1\textwidth]{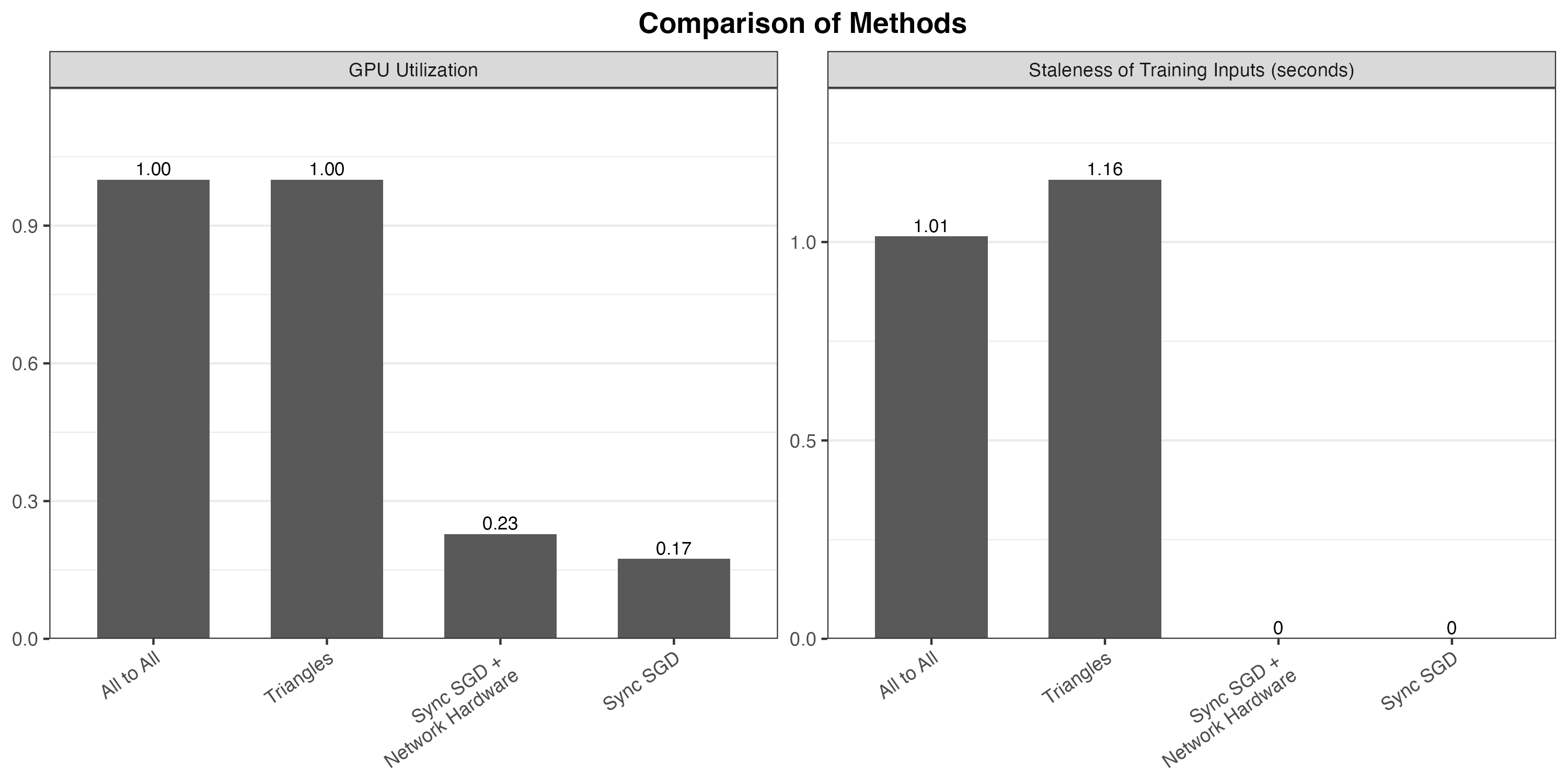}
\caption{GPU utilization and staleness, where staleness is the objective value from Equation \eqref{eq:objective}, across four training systems.}
\label{fig:comparison}
\end{figure}

Figure \ref{fig:comparison} shows the results. It illustrates the tradeoff that our algorithm, like most distributed training algorithms, faces. In synchronous SGD, there is no staleness. No island ever trains on stale parameters. But this comes at a huge cost, which is that the GPUs must idle 80\% of the time in order to support this. Put another way, for tolerating a second of parameter staleness, the system herein can train five times faster. Indeed, most modern distributed training systems (e.g. \citet{douillard2025streaming}) train and synchronize in tandem for this reason. Separately, it is also worth noting that within the two synchronous SGD methods, the hardware certainly helps. The more efficient aggregation and replication that FPGAs and multicast afford allows exchanges to proceed faster, which lowers GPU idling time.

\section{Conclusion}
\label{sec:conclusion}
It is becoming increasingly clear that the global demand for models is outstripping the ability of individual data centers to generate them. This will make distributed training more important than ever, and we believe that making the network an active component of training is the key to the next generation of innovations. While \citet{hoefler2026incollective} and others have voiced skepticism about such approaches, we believe such skepticism better applies to the cases where the infrastructure is limited and shared. Private programmable WANs that can offer bespoke control, fast compute, and ample memory are entirely different and far more powerful.

For such a network, our paper offers two innovations. On the systems front, it uses multicast and FPGA technologies across the WAN to efficiently replicate and aggregate information. On the algorithms front, it creates an optimization framework to turn the WAN's properties into an optimal synchronization schedule. The two are built for each other, and our simulations demonstrate how they couple to offer a self-contained design for distributed training.

Thus far, the paper is primarily theoretical. The DoubleZero network, while a live network that carries traffic of multiple tenants, does not connect to enough clusters of GPUs to train a model in earnest, and compare against its peer systems. This is the next step in our broader push to close the gap between distributed training and its colocated gold standard.

\newpage
\bibliography{refs}

\newpage
\appendix
\section{Model Extensions}
The framework in Section \ref{sec:algorithm} is simple and mostly symmetric, with uniform round lengths, homogeneous stream rates per clique, information exchanges that must complete in their rounds, uniform targets, etc. These were done in the name of a tractable model. But many of them can and should be relaxed in a real-world deployment, as they create inefficiencies. We enumerate them here in three broad categories without implementing them.

\subsection{Time Efficiency}
The core model has a simple notion of time: every round takes $T$ seconds and every exchange must complete within its round. This can be relaxed in a few ways. 
\begin{enumerate}
\item \textbf{Varying round lengths}: for simplicity and tractability, every round in the core model is set to the same length $T$. In practice, nothing requires this, and one could extend the model to have varying round lengths. For instance, one interesting schedule alternates mixing adjacent islands (e.g. islands on the same continent) and distant islands (e.g. islands across continents); and this is a viable schedule under this relaxation, with short rounds handling the former and long rounds handling the latter. The objective function generalizes easily, as the loss in Equation \eqref{eq:objective} simply becomes duration-weighted. The constraint Equation \eqref{eq:completionconstraint} now applies per round type, and so the linear program in Equation \eqref{eq:cliquelp} must be specified and solved per round type individually.
\item \textbf{Overlapping rounds}: in the core model, every movement of data must complete by the round's end, such that no data from round $k$ is in the network once data from round $k + 1$ begins to travel. But in practice, this is unnecessarily tight. For instance, an FPGA can theoretically accept inbound data from round $k + 1$ while draining data from round $k$. Indeed, \citet{douillard2025streaming} generalizes the original DiLoCo algorithm into Streaming DiLoCo for similar reasons. The simple model in this paper does not lend itself to such micro-scheduling, but it can be simulated and operationalized in practice regardless.
\end{enumerate}

\subsection{Network Efficiency}
The core model has an equally simple view of the network. Network routing is always the shortest path, and any links used by multiple routes split that capacity equally. This is potentially inefficient, because resources can be conserved through alternate forms of routing and potentially even given to the slowest clique (which is the binding constraint). As such, this can be relaxed in a few ways.
\begin{enumerate}
\item \textbf{Per-sender stream rates}: in the baseline model, stream rates are held fixed at the clique level. But rates can be set at the sender level (per clique) to potentially ease the memory constraint. More specifically, in asymmetrically spaced cliques, memory has to hold nearby islands' contributions while waiting for those of faraway islands to arrive. With per-sender stream rates, those nearby islands can stream more slowly and lower the memory burden. In practice, this requires a more precise model of byte-level alignment in the FPGAs.
\item \textbf{Intelligent capacity sharing}: for simplicity, any routes that share a link share that capacity equally. That is a strong simplification, for certain cliques may not use all allocated capacity while other cliques are starved for it. Indeed, \citet{gebara2021panama} implement a form of sharing bandwidth between concurrent flows within a data center. While straightforward conceptually, this does turn the linear program in Equation \eqref{eq:cliquelp} into a convex program.
\item \textbf{Flexible routing}: the shortest-path algorithm is the natural default, but there are two reasons why some cliques may want to deliberately add latency and take more circuitous paths through the network. First, such routing potentially frees up capacity on core links for other cliques (especially ones that are the binding constraint). Second, longer routes can restore symmetry to a clique (or, more formally, path delays are sender-receiver delays whereas the core formulation only uses sender delays), which eases the memory utilization. Conceptually, fiber glass here functions as a form of memory, and so it holds the bytes rather than an FPGA holding the bytes. \citet{liu2022optical} for instance reconfigure the optical wide-area topology itself to construct more favorable paths for distributed training. However, this turns the linear program in Equation \eqref{eq:cliquelp} into a pure combinatorial program, which limits its scalability.
\item \textbf{FPGA placement}: FPGAs in this setup are placed on the edges of the network, positioned at the final pathway from the network to each island. This allows them to perform aggregation before this last hop, but in theory they could be placed within the center of the network to condense data streams passing through them on the fly. This relaxes the hardware model itself, and so would change the decision variables. It also requires a very fine-grained model of data movement.
\item \textbf{Resilience}: in this model, there is no jitter in the link latencies, there are no outages in the network, there are no failures at the FPGA, etc. This of course is different in practice, and so any real-world application of the model in this paper has to account for some tolerance that might impair efficiency but enhance robustness. \citet{keiblinger2025pccl} demonstrate the required machinery in practice, with a collective communications library that can tolerate islands joining, leaving, and failing mid-run.
\end{enumerate}

\subsection{Parameter Exchange}
The paper's contribution is the network layer, but there is also potential to make training more efficient or robust at the parameter exchange level. For completeness, we briefly enumerate some relaxations in this domain.
\begin{enumerate}
\item \textbf{Smaller payloads}: the payload $D$ is treated as a constant, but in practice the sender only needs to send parameters that the receivers lack. This means that islands could difference state against some earlier common benchmark and ship a smaller delta, which eases feasibility thresholds for schedules. Indeed, such deltas are the starting place for the compression literature, e.g. \citet{alistarh2017qsgd} and \citet{lin2018dgc}. Moreover, this can be generalized to cases where there is no common benchmark, which more directly trades off payload size versus mixing quality.
\item \textbf{Non-uniform weights}: Equation \eqref{eq:blockmatrix} can be generalized to any doubly-stochastic matrix, i.e. rows and columns summing to one. This can be useful for history-aware weights. As one example, an island receiving updates from two islands that share common history and one with independent history should upweight that last island's contribution relative to the first two.
\item \textbf{Non-uniform targets}: in the core model, all islands are assumed to be homogeneous when in fact they could have heterogeneous compute or data. This means the target weight in the loss function might be an arbitrary $\pi$ rather than specifically $1/N$, or that there might be a more bespoke mapping of distance to aggregate loss rather than the L1 norm across all islands.
\item \textbf{Forgetting}: in the core model, updates accumulate without decay. In practice, real training will partially forget old updates, and so a discount $\rho$ could slot into the area function to better align the loss function with real-world training.
\end{enumerate}
\end{document}